\documentclass[letterpaper, 10 pt, journal, twoside]{IEEEtran}

\usepackage{graphicx}
\usepackage{graphicx, subfig}
\usepackage{amssymb}
\usepackage{hyperref}
\usepackage{float}
\usepackage{booktabs}
\usepackage{multirow}
\usepackage{cite}
\usepackage{amsmath}
\usepackage{color}
\usepackage[dvipsnames]{xcolor}

\newcommand{\HL}[1]{\textcolor[rgb]{1.00,0.00,0.00}{#1}}
\newcommand{\BL}[1]{\textcolor[rgb]{0.00,0.00,1.00}{#1}}
\newcommand{\GL}[1]{\textcolor[rgb]{0.00,0.4,0.00}{#1}}

\begin{document}

\title{\LARGE \bf
Multi-Correlation Siamese Transformer Network with Dense Connection for 3D Single  Object Tracking
}

\author{Shihao~Feng$^*$, Pengpeng~Liang$^{*\dagger}$, Jin Gao, Erkang Cheng%

\thanks{Shihao Feng and Pengpeng Liang are with School of Computer and Artificial Intelligence, Zhengzhou University, Zhengzhou 450001, China (e-mail:  shihaof.0120@qq.com, liangpcs@gmail.com).}%
\thanks{Jin Gao is with State Key Laboratory of Multimodal Artificial Intelligence Systems, Institute of Automation, Chinese Academy of Sciences, Beijing 100190, China, and also with School of Artificial Intelligence, University of Chinese Academy of Sciences, Beijing 100190, China (e-mail: jin.gao@nlpr.ia.ac.cn)}
\thanks{Erkang Cheng is with Nullmax Inc, Shanghai, 201210, China (e-mail: twokang.cheng@gmail.com)}
\thanks{Digital Object Identifier (DOI): 10.1109/LRA.2023.3325715}
\thanks{$^*$Equal contributions. }
\thanks{$^{\dagger}$ Corresponding author. }}

\markboth{IEEE Robotics and Automation Letters. Preprint Version. Accepted October, 2023}
{Feng \MakeLowercase{\textit{et al.}}: Multi-Correlation Siamese Transformer Network with Dense Connection for 3D Single  Object Tracking} 


\maketitle

\begin{abstract}
Point cloud-based 3D object tracking is an important task in autonomous driving. Though great advances regarding Siamese-based 3D tracking have been made recently, it remains challenging to learn the correlation between the template  and  search  branches  effectively with the sparse LIDAR point cloud data. Instead of performing correlation of the two branches at just one point in the network, in this paper, we present a multi-correlation Siamese Transformer  network that has multiple stages and carries out feature correlation at the end of each stage based on sparse pillars. More specifically, in each stage, self-attention is first applied to each branch separately to capture the non-local context information.  Then,  cross-attention is used to inject the  template information into  the search area.  This strategy allows the feature learning of the search area to be aware of the template while keeping the individual characteristics of the template intact.  To enable the network to easily preserve the information learned at different stages  and ease the optimization, for the search area, we densely connect the initial input sparse pillars and the output of each stage to all subsequent stages and the target localization network, which converts pillars to  bird’s eye view (BEV) feature maps and predicts the state of the target with a small densely connected convolution network. Deep supervision is added to each stage to further boost the performance as well. The proposed algorithm is evaluated on the popular KITTI,  nuScenes, and Waymo datasets, and the experimental results show that our method achieves promising performance compared with the state-of-the-art. Ablation study that shows the effectiveness of each component is provided as well.  Code is available at \url{https://github.com/liangp/MCSTN-3DSOT}.
\end{abstract}

\hspace*{\fill} \\
\begin{IEEEkeywords}
3D object tracking, Point cloud, Transformer 
\end{IEEEkeywords}

\section{INTRODUCTION}

\IEEEPARstart{G}{iven} the  3D bounding box of the object in the first frame, the goal of 3D single object tracking (SOT) is to estimate its 3D state in subsequent frames, and it  plays an important role in autonomous driving~\cite{luo2018fast}.  With the development and decreasing cost of the LIDAR sensor,  3D SOT in point cloud has attracted much attention recently~\cite{cui2022exploiting,wu2022multi,V2B,STNet,BAT,mm-track}. However, due to the sparsity and irregularity of the point cloud generated by LIDAR, 3D SOT is still quite challenging.

\begin{figure}[!t]
	\centering
	\includegraphics[width=3.5in]{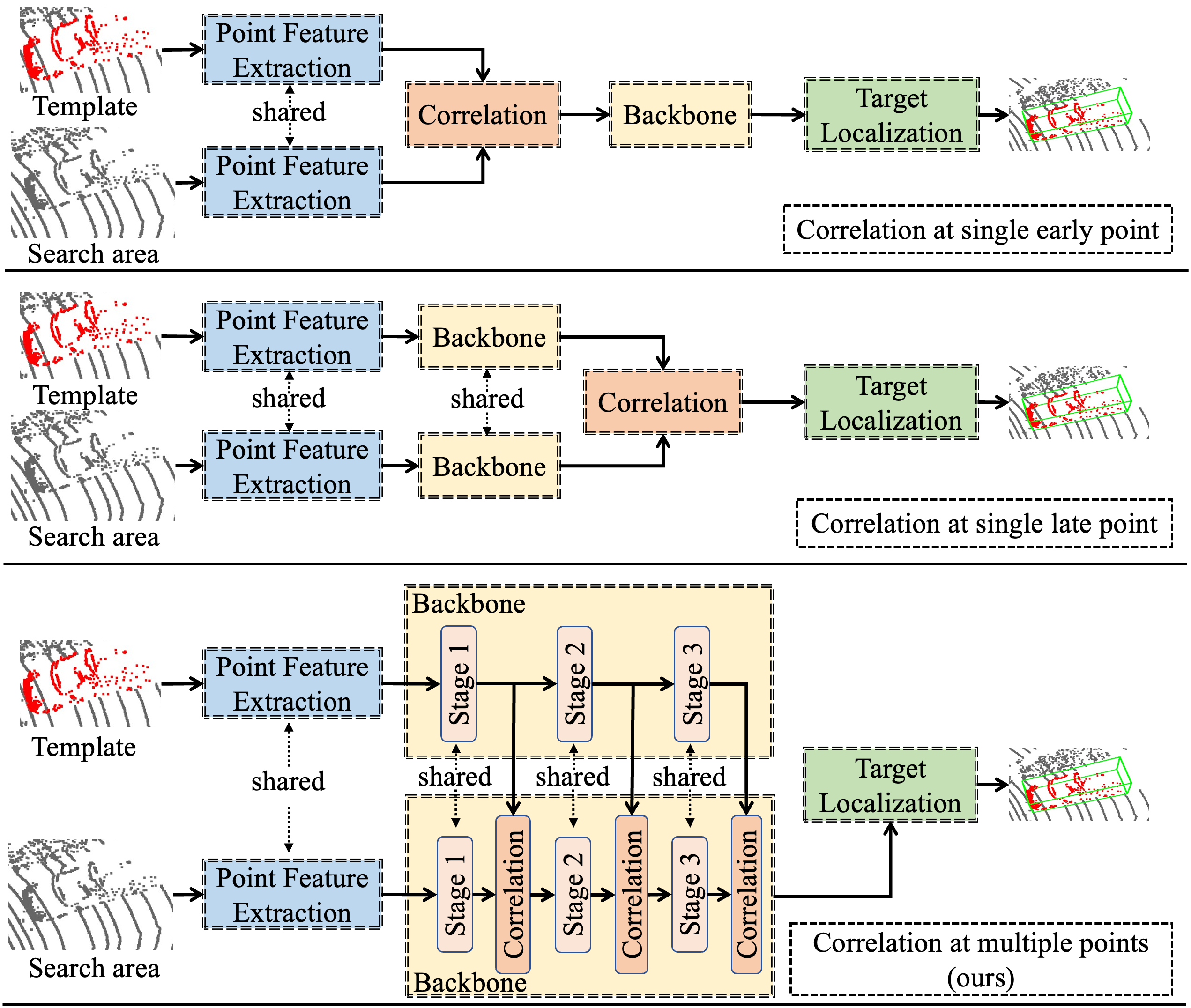}
	\caption{Comparison of different approaches to correlating template and search region.}
	\label{fig:fusion-compare}
\end{figure}

Inspired by the success of the Siamese-based approach in 2D SOT~\cite{bertinetto2016fully,li2018high,chen2021transformer}, most of the recent 3D SOT algorithms~\cite{cui2022exploiting,wu2022multi,SC3D,P2B,V2B,STNet,BAT,cui20213dbmvc} adopt the Siamese framework as well.  Siamese-based trackers first use the initial state of the object to build a template, then correlation learning is performed to fuse the features of the template and the search region around the previous center of the object, and the target is located with the fused features at last. Correlation learning injects the template information into the search region and is a critical step of Siamese-based trackers, and several recent works have made efforts to improve the correlation step for  3D SOT by using point-wise similarity~\cite{P2B,V2B}, adopting Transformer~\cite{cui2022exploiting,cui20213dbmvc,STNet}, and leveraging box information~\cite{BAT}.  


However, the above approaches just fuse the template and search branches at only one point in the network. Based on where the correlation happens, they either sacrifice the individual semantic information of the template or are difficult to learn the representation of the search  region with sufficient awareness of the template. On the one hand, if the correlation is located at an early point of the network (top of Fig.~\ref{fig:fusion-compare}), the intact semantic information of the template cannot be maintained. On the other hand, if the correlation is situated at a late point of the network (middle of Fig.~\ref{fig:fusion-compare}), the feature extraction of the search region  lacks template information. We conjecture that both keeping the template branch intact and making the search region aware of the template during the feature extraction process can benefit the tracking performance.


In this paper, we propose to learn the representations of the template and search region separately until the target localization network, and the template information is injected into the search branch at multiple stages of the network (bottom of Fig.~\ref{fig:fusion-compare}). More specifically,  taking inspiration from the recently proposed SST architecture~\cite{fan2022embracing}, given the point cloud of the template and the search region, we first use a simplified PointNet~\cite{qi2017pointnet} to extract point features and aggregate them into sparse pillars for each branch. Then, a multi-correlation Siamese Transformer network with multiple stages is designed to process the sparse pillars of the two branches. In each stage, the two branches are processed with self-attention individually, and cross-attention is used to fuse information from template to search region.  To further boost the performance, for the search area, we connect the features of the initial sparse pillars and the output of each stage to the inputs of all subsequent stages and the final target localization network, and this strengthens the propagation and preservation of the features learned at each stage  and makes the optimization easier~\cite{densenet}. The target localization network converts the input sparse pillars to bird’s eye view (BEV) feature maps, and a small densely connected convolution block is adopted to predict the state of the target.  Besides, we add deep supervision to each stage to ease the optimization further. We carry out extensive experiments on the KITTI~\cite{kiitidataset}, nuScenes~\cite{nuscenesdataset}, and Waymo~\cite{waymodataset} datasets, and the proposed algorithm achieves promising results. 

Our main contributions are summarized as follows:

\begin{itemize}
 \item We propose to inject the intact template information into the search area at multiple points of the network so that the feature extraction of the search area is aware of the template for Siamese-based 3D SOT.
 \item We design a multi-stage densely connected Siamese Transformer network that uses self-attention for separate feature learning of each branch and cross-attention for feature correlation between the template and search area based on sparse pillars in each stage.
 \item Comprehensive experiments on the KITTI~\cite{kiitidataset}, nuScenes~\cite{nuscenesdataset}, and Waymo~\cite{waymodataset} show the performance of the proposed algorithm is promising in comparison with the state-of-the-art algorithms. 
 \end{itemize}

\begin{figure*}[!t]
	\centering
	\vspace*{0.1cm}
	\includegraphics[width=\linewidth]{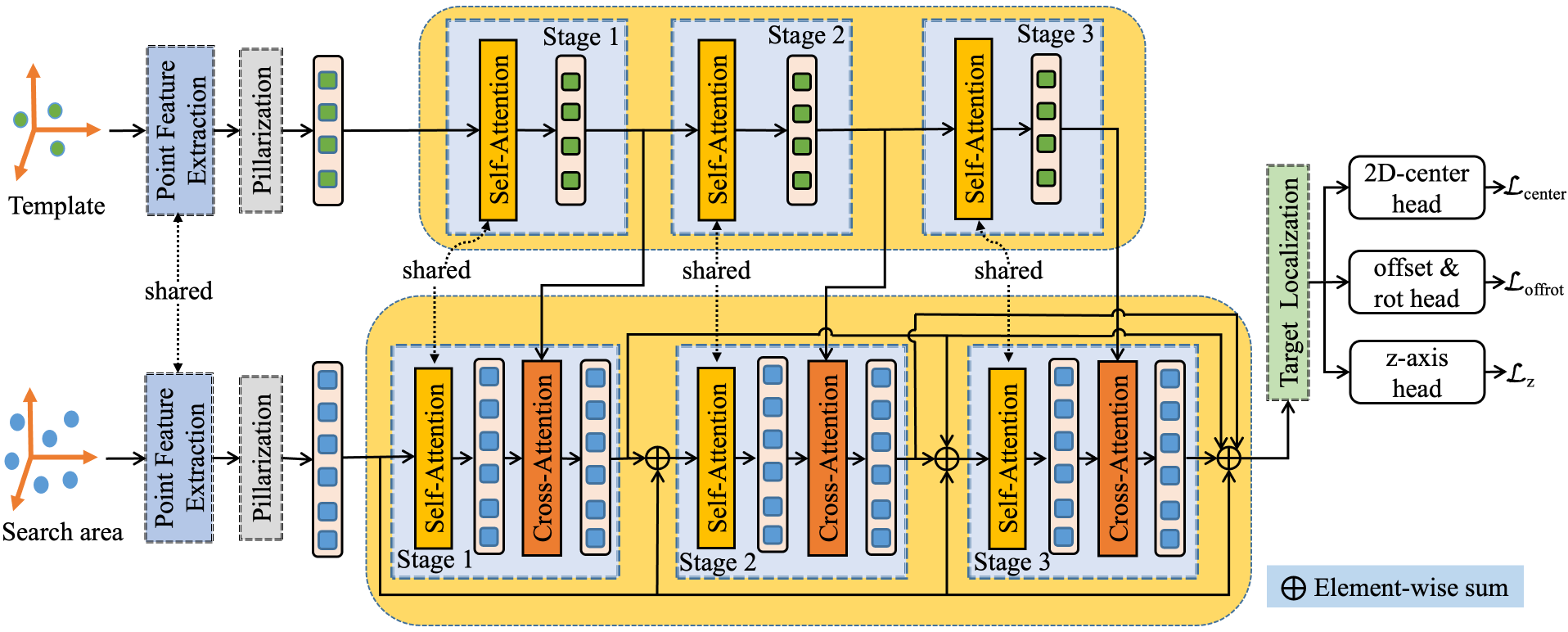}
	\caption{Overall framework of our approach. Point features of the input point clouds of the template and search region are extracted and converted to pillars first.  The pillars are further processed by a multi-stage (three stages are used here for illustration purpose) Siamese network with self-attention to extract features, and cross-attention is used to fuse the two branches in each sage. Lastly, a target localization network converts the learned representations of the pillars of the search region to BEV feature maps and predicts the state of the target.}
	\label{fig:framework}
\end{figure*}

\section{Related Work}
\noindent\textbf{2D Siamese Tracking.}  Siamese-based approaches~\cite{bertinetto2016fully,li2018high,chen2021transformer,zhang2020ocean,li2019siamrpn++,lin2022swintrack,cui2022mixformer} are  mainstays  for 2D SOT in the recent several years, and they predict the location and size of the object based on the similarity between the features of the template and search region.  The pioneering work SiamFC~\cite{bertinetto2016fully} uses a Siamese fully-convolutional network to extract feature maps of the template and search region, and it employs a correlation layer to calculate the similarity score map between the two branches. SiamRPN~\cite{li2018high} extracts a set of proposals with the correlation feature maps, and proposal selection strategies are designed to locate the target.    ~\cite{zhang2020ocean} presents an anchor-free tracker, and every pixel in the ground truth bounding box can contribute to training.  \cite{chen2021transformer} exploits Transformer~\cite{vaswani2017attention}  to fuse the template and search features.~\cite{lin2022swintrack} adapts Swin Transformer~\cite{liu2021swin} to tracking and proposes to use a motion token to embed the historical target trajectory. In ~\cite{cui2022mixformer}, Transformer is used  to extract image features and fuse the two branches simultaneously without using CNN for feature extraction first. It is worth noting our work is distinct from~\cite{cui2022mixformer} by designing a network specifically for LIDAR point cloud and keeping the feature extraction of the template separate from the search region.  


\noindent\textbf{3D Single Object Tracking.} Before LIDAR sensors become popular, 3D single object tracking algorithms~\cite{pieropan2015robust,kart2019object,song2013tracking} mainly use RGB-D data, and they heavily rely on RGB information.  As a pioneering work focusing on 3D SOT on LIDAR point cloud, SC3D~\cite{SC3D} proposes to learn a Siamese network with shape completion regularization for measuring the similarity between the template and search candidates. Two shortcomings of SC3D are low speed caused by a large number of search candidates and not allowing end-to-end training. To solve these two problems, P2B~\cite{P2B} augments the representation of the search branch with template point features and generates 3D target proposals for verification via voting and clustering.    ~\cite{cui2022exploiting} first converts the point cloud to dense multi-scale features via BEV, then it uses Transformer to capture global information and fuse the template and search region. BAT~\cite{BAT} proposes to represent the object via the relation between the points and the corners and center of the box so that the size and part information can be captured.  ~\cite{wu2022multi} leverages graph to capture the structural information embedded in the point cloud, and a cross-graph network is designed for the fusion of the target and search branches. Instead of relying on appearance matching, in~\cite{mm-track}, the points of two consecutive frames with additional segmentation labels are used to estimate a coarse motion of the target, then the motion is refined with a denser point cloud by aggregating two partial target views. ~\cite{shan2021ptt} uses self-attention and position encoding to consider the varying importance of different points.   \cite{wang2021mlvsnet} proposes to perform Hough voting with a designed target-guided attention module  at multiple levels to  generate more vote centers.  PTTR~\cite{zhou2022pttr} presents a point relation transformer that can capture long-range dependencies and fuse the template and the search area.

\noindent\textbf{Fusion with Transformer for 3D Object Detection.} LIDAR and camera are two popular complementary types of sensors for 3D object detection, LIDAR can capture sparse depth information while camera can collect dense texture information. As the attention mechanism of Transformer~\cite{vaswani2017attention} allows tokens to aggregate information from each other, Transformer is suitable for feature fusion, and researchers start to explore it to fuse features from these two modalities~\cite{li2022deepfusion,bai2022transfusion,yang2022deepinteraction,autoalignchen2022,chen2022autoalignv2}.~\cite{li2022deepfusion} first reverses the data augmentation on point cloud,  and then transforms the voxels of the point cloud into queries and image features into keys and values, and cross-attention is used for fusion. A two-layer Transformer decoder is proposed in~\cite{bai2022transfusion}, the first layer predicts an initial set of objects and updates the object queries, and the second layer uses the updated queries to fuse image features and refines the detection results.  In~\cite{yang2022deepinteraction}, a separate encoder is maintained for each modality to keep its unique characteristics, and the interaction between two modalities is carried out with cross-attention during encoding. Then, a predictive interaction decoder is designed to make the prediction of one modality depend on the output of another.~\cite{autoalignchen2022} uses cross-attention to aggregate image features into voxels of point cloud. ~\cite{chen2022autoalignv2} improves~\cite{autoalignchen2022} with deformable attention and decomposing the token representation into  domain-specific and instance-specific parts. Besides fusing information from different modalities, ~\cite{tong2023multi} adopts Transformer to fuse features of point cloud represented in different forms, i.e. point, voxel, and BEV.    

%

\section{Method}

\subsection{Overall Architecture}
\label{sec-overall}
In 3D single object tracking on point cloud, the state of an object is represented as $(x,y,z,w,l,h,\theta)\in \mathbb{R}^{7}$, where $(x,y,z)$ is the center of the object, $(w,l,h)$ is the size (width, length, height), and $\theta$ is the yaw angle.  Given the initial object state and the associated template point cloud $P^{t}=\{p_i^{t}\}_{i=1}^{N_t}$, the goal of 3D SOT is to predict the state of the object with the point cloud of a search region $P^{s}=\{p_i^{s}\}_{i=1}^{N_s}$ frame by frame, where $N_t$ and $N_s$ are the numbers of points in the template and search region, respectively.  As the 3D size of the object does not change or changes very little (e.g. pedestrian) across all frames, we just estimate $(x,y,z,\theta)\in \mathbb{R}^{4}$.   As shown in Fig.~\ref{fig:framework}, we design our tracker following the Siamese-based approach. Having $P^s$ and $P^t$ as input,  we first use a densely connected multi-stage Siamese Transformer network (Section~\ref{sec-mutli-stage}) to extract features of the template and search region. During the feature extraction process, the template branch is kept intact, and its information is injected into the search branch at the end of each stage. Then, the features of the search region are input into the target localization network (Section~\ref{sec-target-localization}) to regress the state of the object.


\subsection{Multi-Correlation Siamese Transformer Network with Dense Connection}
\label{sec-mutli-stage}
\noindent\textbf{Point  cloud to pillars.}  Given the point cloud $P^t$ and  $P^s$ of  the template and  search region, respectively, we follow~\cite{pointpillars,fan2022embracing} to convert the point cloud of each branch to sparse pillars. For each branch, each point in the point cloud is projected via dynamic voxelization~\cite{zhou2020end} to a cell of an evenly spaced grid of size $X\times Y$  on the $x$-$y$ plane.  Each cell corresponds to a pillar, and a set of pillars $V$ with $|V|=X\times Y$ is created.  Besides the coordinate $(x,y,z)$, each point in a pillar is further decorated with $x_c,y_c,z_c,x_p,y_p,z_p$, where the subscript $c$ indicates the distance to the mean of the coordinates of all points in the pillar, and the subscript $p$ denotes the offset from the center of the pillar. As a result, the initial representation of a point is $D=9$ dimensional.  Each point is further embedded into a $C$ dimensional feature vector via a simplified PointNet~\cite{qi2017pointnet}, which consists of a linear layer,  BatchNorm, and ReLU. The representation of each non-empty pillar is generated via max pooling over the features of points in it.

\noindent\textbf{Feature learning and correlation.}  The main part of our feature learning and correlation network contains multiple stages as shown in Fig.~\ref{fig:framework}. The input of each stage contains the features of  two sets of non-empty pillars $F^t\in \mathbb{R}^{M_t\times C}$ and  $F^s\in \mathbb{R}^{M_s\times C}$ for the template branch and search branch, respectively, where $M^t$ and $M^s$ are the numbers of non-empty pillars. To capture the non-local context information, we use self-attention to process the non-empty pillars of each branch, and the parameters are shared between the two branches.  More specifically, the coordinates of $M$ pillars are transformed to the positional encoding $E\in \mathbb{R}^{M\times C}$ with two linear layers and ReLU following~\cite{STNet} first. Next, the sum $F+E$ is projected to \emph{query}, \emph{key}, and \emph{value} by three matrices $W_Q^{sa}\in \mathbb{R}^{C\times C}$, $W_K^{sa}\in \mathbb{R}^{C\times C}$, and $W_V^{sa}\in \mathbb{R}^{C\times C}$, respectively. The processing of $F^t$ and $F^s$ is defined as follows:
\begin{equation}
	\hat{F}^{t} = SA(W_Q^{sa}(F^t+E^t),W_K^{sa}(F^t+E^t),W_V^{sa}(F^t+E^t))
\end{equation}
\vspace{-6mm}
\begin{equation}
\label{eq:self-attention-search}
	\hat{F}^{s} = SA(W_Q^{sa}(F^s+E^s),W_K^{sa}(F^s+E^s),W_V^{sa}(F^s+E^s))
\end{equation}
where $SA$ (short for self-attention) is the linear attention~\cite{katharopoulos2020transformers}, which is used by~\cite{STNet} as well. 

After processing the non-empty pillars of the two branches, to make the feature learning of the search branch aware of the target information, at the end of each stage, we use cross-attention to inject the template information into the search branch. As the target localization network (Section~\ref{sec-target-localization}) takes the features of the search region as input,  we add the positional encoding $E^s$ to $\hat{F}^s$ so that the position information in the search region can be enhanced in the feature fusion process, and the sum  $\hat{F}^s+E^s$ is projected to \emph{query} by $W_Q^{ca}\in \mathbb{R}^{C\times C}$,  $\hat{F}^t$ is projected to \emph{key}  and ~\emph{value} by $W_K^{ca}\in \mathbb{R}^{C\times C}$ and $W_V^{ca}\in \mathbb{R}^{C\times C}$, respectively. We also use linear attention~\cite{katharopoulos2020transformers} for  $CA$ (short for cross-attention) below:
\begin{equation}
\label{eq:cross-attention}
	\tilde{F}^{s} = CA(W_Q^{ca}(\hat{F}^s+E^s),W_K^{ca}(\hat{F}^t),W_V^{ca}(\hat{F}^t))
\end{equation}
$\hat{F}^{t}$ and $\tilde{F}^s$ are fed to the next stage for further processing, and the template is kept intact during the entire feature learning process.

\noindent\textbf{Densely connected stages. }  To ease the information flow across stages, inspired by~\cite{densenet}, for the search branch, we densely connect the input features of pillars and the output of each stage to the inputs of all subsequent stages via element-wise sum as shown in Fig.~\ref{fig:framework}. In each stage, the search branch receives all the preceding features, which are summed as input:
\begin{equation}
	F^{s}_{i} = \sum_{l=0}^{i-1}\tilde{F}^{s}_{l}
\end{equation}
where $F^{s}_{i}$ is the input of the self-attention in Eq.~\ref{eq:self-attention-search} at stage $i$, $\tilde{F}^s_l$ is the output of the cross-attention in Eq.~\ref{eq:cross-attention} of stage $l$ for $l>0$, and  $\tilde{F}^s_l$ is the initial pillar features of the  search region when $l=0$. 

\subsection{Target Localization Network}
\label{sec-target-localization}
 The sum of the initial sparse pillars and the outputs of all stages of the search area is used as the input for target localization, and we build the target localization network on top of BEV features based on~\cite{V2B}.  Following~\cite{fan2022embracing}, the feature of each pillar in $\tilde{F}^{s}_{loc}$ is placed back to its corresponding BEV grid in the $x$-$y$ plane, and the unfilled grids are set to zeros. As the object center can be located in an empty zero grid, the BEV feature maps are further processed with three $3\times 3$ convolutions, which can fill most of the zero holes. These three convolutions are densely connected by element-wise adding the initial input and the output of  each convolution to the inputs of all subsequent convolutions.  The detection head is the same as~\cite{V2B}, and it includes three parts: (1) \emph{2D center head} that  predicts the discrete integer 2D center in the $x$-$y$ plane, (2) \emph{offset and rotation head} that regresses  the offset from the continuous floating-point 2D center for the continuous ground truth center and a rotation angle (yaw), and (3)  \emph{$z$-axis head} that estimates the location along $z$-axis.  The loss function is defined as follows:
\begin{equation}\label{eq:loss}
	\mathcal{L} = \lambda_{1}(\mathcal{L}_{center} + \mathcal{L}_{offrot}) + \lambda_2\mathcal{L}_{z}
\end{equation}
where $\mathcal{L}_{center}$ is the focal loss~\cite{lin2017focal}, $\mathcal{L}_{offrot}$ and $\mathcal{L}_{z}$ are the $L_1$ loss. 

Besides $\mathcal{L}$, to further reduce the optimization difficulty, inspired by previous works~\cite{xie2015holistically}, during the training phase, we convert the output pillars at the end of each stage to BEV feature maps in the same way as above and add deep supervision in each stage. The final loss function is as follows:
\begin{equation}\label{eq:finalloss}
	\mathcal{L}_{final} = \mathcal{L} + \alpha\sum_{i=1}^{S-1}\mathcal{L}_{i}
\end{equation}
where $\mathcal{L}_i$ is calculated in the same way as $\mathcal{L}$, and $S$ is the total number of stages. For the experiment, we set $\lambda_{1}=1.0$, $\lambda_2=2.0$, and $\alpha=0.1$.

\section{Experiments}

\begin{table*}[htb]
	\caption{Comparison with state-of-the-art algorithms on the KITTI dataset. The best three results are indicated by \HL{red}, \BL{blue}, and \GL{green}, respectively. ``MC", ``FFS" and ``RS" are short for  ``motion centric", ``full frame-based similarity" and ``region-based similarity", respectively. The number after each category is the total number of frames for test.}
	\label{table:kitti-result}
	\centering
	\begin{tabular}{c|c|cc|cc|cc|cc|cc}
		\toprule[.05cm]
		\multirow{2}{*}{Method} & \multirow{2}{*}{Paradigm} & \multicolumn{2}{c|}{Car-6424} & \multicolumn{2}{c|}{Pedestrian-6088} & \multicolumn{2}{c|}{Van-1248} & \multicolumn{2}{c|}{Cyclist-308} & \multicolumn{2}{c}{Mean-14068} \\
		&                           & Success    & Precision   & Success    & Precision   & Success    & Precision   & Success    & Precision   & Success    & Precision   \\ 
		\midrule
		{$M^2$-Track~\cite{mm-track}} & MC   & {65.5}      & {80.8}        & {\textcolor{blue}{61.5}}      & {\textcolor{blue}{88.2}}       & {53.8}       & {\GL{70.7}}        & {\GL{73.2}} &  {\GL{93.5}}  & {\GL{62.9}} & {\textcolor{blue}{83.4}}        \\ 
		
		\midrule
		{CXTrack\cite{CXTrack}} & FFS   & {69.1}      & {81.6}        & {\textcolor{red}{67.0}}      & {\textcolor{red}{91.5}}       & {\textcolor{blue}{60.0}}       & {\textcolor{blue}{71.8}}        & {\textcolor{red}{74.2}} &  {\textcolor{red}{94.3}}  & {\textcolor{red}{67.5}} & {\textcolor{red}{85.3}}        \\ 
		
		\midrule
		SC3D~\cite{SC3D}      & \multirow{10}{*}{RS}                & 41.3       & 57.9        & 18.2       & 37.8        & 40.4       & 47.0        & 41.5       & 70.4        & 31.2       & 48.5        \\
		P2B~\cite{P2B}      &               & 56.2       & 72.8        &28.7        & 49.6        & 40.8       & 48.4        & 32.1       & 44.7        & 42.4       & 60.0        \\
		BAT~\cite{BAT}        &                 & 60.5       & 77.7        & 42.1       & 70.1        & 52.4       & {67.0}       & 33.7       & 45.4        & 51.2       & 72.8        \\
		V2B~\cite{V2B}       &   & {70.5}       & 81.3        & 48.3       & 73.5        & 50.1       & 58.0        & 40.8       & 49.7        & 58.4       & 75.2        \\
		SMAT~\cite{cui2022exploiting}       &    & \GL{71.9} & \GL{82.4} & {52.1} & 81.5        & 41.4       & 53.2        & 61.2       & 87.3        & {60.4}       & {79.5}        \\
		STNet~\cite{STNet}       &   & \textcolor{blue}{72.1}       & \textcolor{blue}{84.0}        & 49.9       & 77.2        & \GL{58.0}       & 70.6        & \textcolor{blue}{73.5}       & \textcolor{blue}{93.7}        & 61.3       & 80.1        \\ 
		OSP2B~\cite{nie2023osp2b}       &   & {67.5}       & 82.3        & 53.6       & \GL{85.1}        & 56.3       & 66.2        & 65.6       & 90.5        & 60.5       & 82.3        \\
		MLSENet~\cite{wu2022multi}       &   & {69.7}       & 81.0        & 50.7       & 80.0        & 55.2       & 64.8        & 41.0       & 49.7        & 59.6       & 78.4        \\
		Ours       &   & \textcolor{red}{73.6}       & \textcolor{red}{84.7}        & \GL{56.8}       & 83.7       & \textcolor{red}{62.6}       & \textcolor{red}{74.4}      & 41.4      & 54.6       & \textcolor{blue}{64.6}       & \GL{82.7}        \\ 
		\bottomrule[.05cm]
	\end{tabular}
\end{table*}

\begin{table*}[htbp]
	\caption{Comparison with state-of-the-art algorithms on the nuScenes dataset. Please refer to Table~\ref{table:kitti-result} for the meanings of the text colors, abbreviations for the paradigm, and the number after each category. }
	\label{table:result-nuScenes}
	\centering
	\begin{tabular}{c|c|cc|cc|cc|cc|cc}
		\toprule[.05cm]
		\multirow{2}{*}{Method} & \multirow{2}{*}{Paradigm} & \multicolumn{2}{c|}{Car-15578} & \multicolumn{2}{c|}{Pedestrian-8019} & \multicolumn{2}{c|}{Truck-3710} & \multicolumn{2}{c|}{Bicycle-501} & \multicolumn{2}{c}{Mean-27808} \\
		 & & Success    & Precision   & Success    & Precision   & Success    & Precision   & Success    & Precision   & Success    & Precision   \\ 
		 \midrule
		 {CXTrack\cite{CXTrack}} & FFS  & 29.6 & 33.4 & \textcolor{red}{20.4} & \textcolor{red}{32.9} & \textcolor{red}{27.6} & \textcolor{blue}{20.8} & 18.5 & \textcolor{blue}{26.8} & \GL{26.5} & \textcolor{red}{31.5}  \\
		
		\midrule
		SC3D~\cite{SC3D} & \multirow{6}{*}{RS}                 & 25.0       & 27.1        & 14.2       & 16.2        & \textcolor{blue}{25.7}       & \textcolor{red}{21.9}        & 17.0       & 18.2        & 21.8       & 23.1        \\
		P2B~\cite{P2B} &     & 27.0      & 29.2     &15.9        & 22.0       & 21.5      & 16.2        & 20.0       & \GL{26.4}        & 22.9       & 25.3       \\
		BAT~\cite{BAT}  &       & 22.5       & 24.1        & 17.3      & {24.5}       & 19.3       & 15.8       & 17.0       & 18.8        & 20.5       & 23.0        \\
		V2B~\cite{V2B}   &    & \GL{31.3}       & \GL{35.1}       & {17.3}       & 23.4        & 21.7       & 16.7        & \textcolor{red}{22.2}      & 19.1        & {25.8}       & {29.0}        \\
		STNet~\cite{STNet} &     & \textcolor{blue}{32.2}     &  \textcolor{blue}{36.1}      & \GL{19.1}     &\GL{27.2}    & {22.3}      & {16.8}       & \textcolor{blue}{21.2}    &  \textcolor{red}{29.2}     & \textcolor{blue}{26.9}     & \GL{30.8}      \\ 
		
		Ours   &   & \textcolor{red}{32.5}  & \textcolor{red}{36.2}   & \textcolor{blue}{19.4}    & \textcolor{blue}{28.3}    &  \GL{22.6}   & \GL{18.0}    & \GL{20.4}      & {19.8}      & \textcolor{red}{27.2}    & \textcolor{blue}{31.2}    \\ 
		\bottomrule[.05cm]
	\end{tabular}
\end{table*}

\subsection{Experimental Settings}

\noindent\textbf{Datasets.} We comprehensively evaluate our method on three popular datasets: KITTI~\cite{kiitidataset}, nuScenes~\cite{nuscenesdataset}, and Waymo Open Dataset (WOD)~\cite{waymodataset}. As these datasets are not originally curated for 3D SOT, we follow~\cite{V2B} to adapt them. More specifically, due to the inaccessibility of the ground truth of test data,  the 21 training video sequences of the KITTI dataset are divided into 0-16 for training, 17-18 for verification, and 19-20 for test. For  the nuScenes dataset,  its 150 validation sequences  are used for evaluation. For WOD,  the selected 1121 tracklets by~\cite{LidarSOT} are categorized into easy, medium, and difficult subsets based on the number of points in each tracklet's first frame. Note that the evaluations on nuScenes and WOD are carried out with the model trained using the KITTI training subset following~\cite{V2B,STNet}. The car, pedestrian, van, and cyclist classes of  KITTI training data contain 19522, 4600, 1994, 1529 frames, respectively.

\noindent\textbf{Evaluation metrics.}  Following~\cite{SC3D,V2B,STNet}, we measure \emph{Success} and \emph{Precision} with one pass evaluation (OPE)~\cite{wu2013online}.  \emph{Success} is defined as the intersection over union (IOU) between the 3D predicted bounding box (BBox) and the ground truth (GT) bounding box. \emph{Precision} is the area under curve (AUC) of a precision plot obtained by varying the distance between the centers of the two bounding boxes from 0 to 2m. 



\noindent\textbf{Implementation details.}  We randomly sample (discarding or duplicating) $N_{t}=512$ and  $N_{s}=1024$ points for the template and search region, respectively, as do~\cite{STNet,P2B}. The implementation of dynamic voxelization~\cite{zhou2020end} from MMDetection3D~\cite{mmdet3d2020}  is adopted to construct pillars.   The grid size of each pillar in the $x$-$y$ plane is set to $(0.3m,0.3m)$, and the feature dimension of each non-empty pillar is 128. The number of stages is set to two premised on the ablation study. After being processed by the multi-stage network, the features of the non-empty pillars are placed back to their corresponding $x$-$y$ grids to build the BEV feature maps. 

\noindent\textbf{Training and test.}  To generate training data, given two adjacent frames of times $t-1$ and $t$, we combine the points in the ground truth bounding box (GTBB) in the first frame and the points in the GTBB with random shift in the $(t-1)$th frame to construct the template; for the search region, the GTBB of the $t$th frame with random shift is enlarged by 2 meters in each direction, and the points within the enlarged box are collected. During test, the template points are made up of the points in the GTBB of the first frame and the points inside the predicted BBox of the previous frame; for the points of search, we enlarge the previous predicted BBox by 2 meters without random shift and use the points inside it. For the optimization of the network, Adam optimizer~\cite{adam} is adopted, and our model is trained with batch size of 32 for 40 epochs in total.

\subsection{Comparison with State-of-the-arts}

\begin{figure*}[htbp]
	\centering
	\vspace*{0.1cm}
	\includegraphics[width=0.9\linewidth]{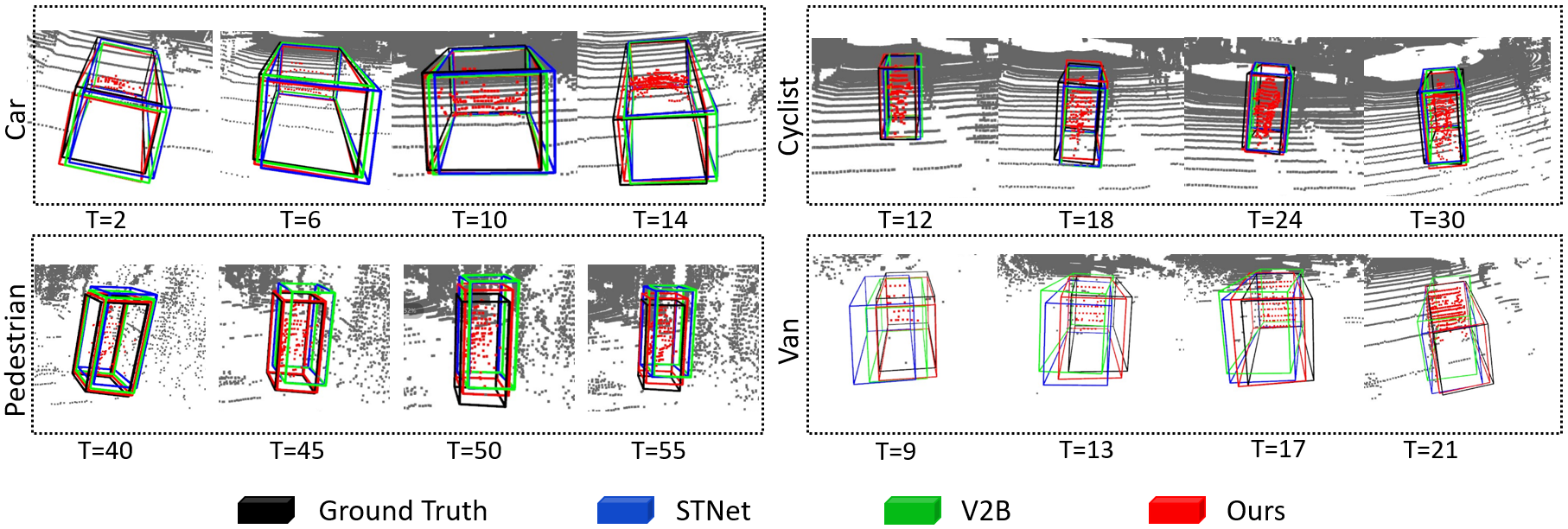}
	\caption{Visualization results of the KITTI dataset.}
	\label{fig:KITTI_visual}
\end{figure*}

\begin{table*}[htbp]
	\caption{Comparison with state-of-the-art algorithms on WOD. Please refer to Table~\ref{table:kitti-result} for the meanings of the text colors and abbreviations for the paradigm. The number under each difficulty level indicates the total number of frames for test.}
	\label{table:result-WOD}
	\centering	
	\begin{tabular}{c|c|c|cccc|cccc}
		\toprule[2pt]
		& \multirow{3}{*}{Method} & \multirow{3}{*}{Paradigm}  & \multicolumn{4}{|c}{Vehicle} & \multicolumn{4}{|c}{Pedestrian} \\
		\cmidrule{4-11}
		 & &  & Easy & Medium & Hard & Mean  & Easy & Medium & Hard & Mean  \\
		&  &  & 67832 & 61252 & 56647 & 185731 & 85280 & 82253 & 74219 & 241752 \\ 
		\midrule
		& {CXTrack\cite{CXTrack}} & FFS & 63.9 & 54.2 & \GL{52.1} & {57.1} & \textcolor{red}{35.4} & \textcolor{red}{29.7} & \textcolor{red}{26.3} & \textcolor{red}{30.7} \\  
		\cmidrule{2-11}
		\multirow{5}{*}{Success}& P2B\cite{P2B} & \multirow{5}{*}{RS} & 57.1 & 52.0 & 47.9 & 52.6  & 18.1 & 17.8 & 17.7 & 17.9 \\
		& BAT\cite{BAT} & & 61.0 & 53.3 & 48.9 & 54.7 & 19.3 & 17.8 & 17.2 & 18.2  \\
		& V2B\cite{V2B} & & \GL{64.5} & \GL{55.1} & {52.0} & \GL{57.6} & {27.9} & {22.5} & {20.1} & {23.7} \\
		& STNet\cite{STNet}&  & \textcolor{blue}{65.9} & \textcolor{blue}{57.5} & \textcolor{blue}{54.6} & \textcolor{blue}{59.7}	&  \GL{29.2} & \GL{24.7} & \GL{22.2} & \GL{25.5} \\
		& ours &  & \textcolor{red}{68.6} & \textcolor{red}{59.3} & \textcolor{red}{55.1} & \textcolor{red}{61.4} & \textcolor{blue}{30.3} & \textcolor{blue}{25.4} & \textcolor{blue}{23.5} & \textcolor{blue}{26.5} \\
		\midrule
		 & {CXTrack\cite{CXTrack}} & FFS & 71.1 & 62.7 & \GL{63.7} & \GL{66.1} & \textcolor{red}{55.3} & \textcolor{red}{47.9} & \textcolor{red}{44.4} & \textcolor{red}{49.4} \\  
		\cmidrule{2-11}
		\multirow{5}{*}{Precision}& P2B\cite{P2B} & \multirow{5}{*}{RS} & 65.4 & 60.7 & 58.5 & 61.7 & 30.8 & 30.0 & 29.3 & 30.1 \\
		& BAT\cite{BAT} & & 68.3 & 60.9 & 57.8 & 62.7 & 32.6 & 29.8 & 28.3 & 30.3  \\
		& V2B\cite{V2B} & & \GL{71.5} & \GL{63.2} & {62.0} & {65.9} & {43.9} & {36.2} & {33.1} & {37.9} \\
		& STNet\cite{STNet} & &\textcolor{blue}{72.7} & \textcolor{blue}{66.0} & \textcolor{blue}{64.7} & \textcolor{blue}{68.0} & \GL{45.3} & \GL{38.2} & \GL{35.8} & \GL{39.9} \\
		& ours & &\textcolor{red}{73.8} & \textcolor{red}{67.6} & \textcolor{red}{65.3} & \textcolor{red}{69.2}  & \textcolor{blue}{48.2} & \textcolor{blue}{39.3} & \textcolor{blue}{36.2} & \textcolor{blue}{41.5} \\
		\bottomrule[2pt]
		
	\end{tabular}
\end{table*}

\noindent\textbf{Results on KITTI.}  We compare our algorithm with ten state-of-the-art algorithms~\cite{SC3D,P2B,wu2022multi,BAT,V2B,cui2022exploiting,STNet,nie2023osp2b,mm-track,CXTrack} on the KITTI dataset~\cite{kiitidataset}. Besides $M^2$-Track~\cite{mm-track} and CXTrack~\cite{CXTrack} which adopt a motion-centric (MC) paradigm and a full frame-based similarity (FFS) paradigm, respectively, our algorithm and all other compared algorithms use a region-based similarity paradigm.  MC  locates the target by directly estimating its motion between two frames. The most recently proposed FFS paradigm in~\cite{CXTrack} tracks the target by modeling the similarity between two full consecutive frames while RS locates the target by calculating the similarity between a cropped  template and a cropped search region. 

As shown in Table~\ref{table:kitti-result}, among the RS approaches, our tracker achieves the best mean performance in terms of both success and precision, and the improvement of success over the second best RS tracker STNet~\cite{STNet} is $3.3\%$  ($64.6\%$ vs $61.3\%$).  The mean performance of our tracker is on par with $M^2$-Track~\cite{mm-track} of MC paradigm, and is $1.7\%$ better than it regarding success while lagging behind it by $0.7\%$ in terms of precision. CXTrack~\cite{CXTrack} using FFS paradigm achieves the best performance across all three paradigms, and this can be attributed to the context information captured by the full frame-based similarity modeling. Meanwhile, our algorithm obtains the best performance on the car and van categories compared with approaches of all three paradigms.  Our algorithm does not perform well in the cyclist category in comparison with the top-ranked trackers, and this  might be caused by the small size of the target (the number of points of the target is small as well) and the less training data compared with the other three categories. Figure~\ref{fig:KITTI_visual} shows some visualized results on the KITTI dataset.    



\begin{figure*}[htbp]
	\centering
	\vspace*{0.1cm}
	\includegraphics[width=0.9\linewidth]{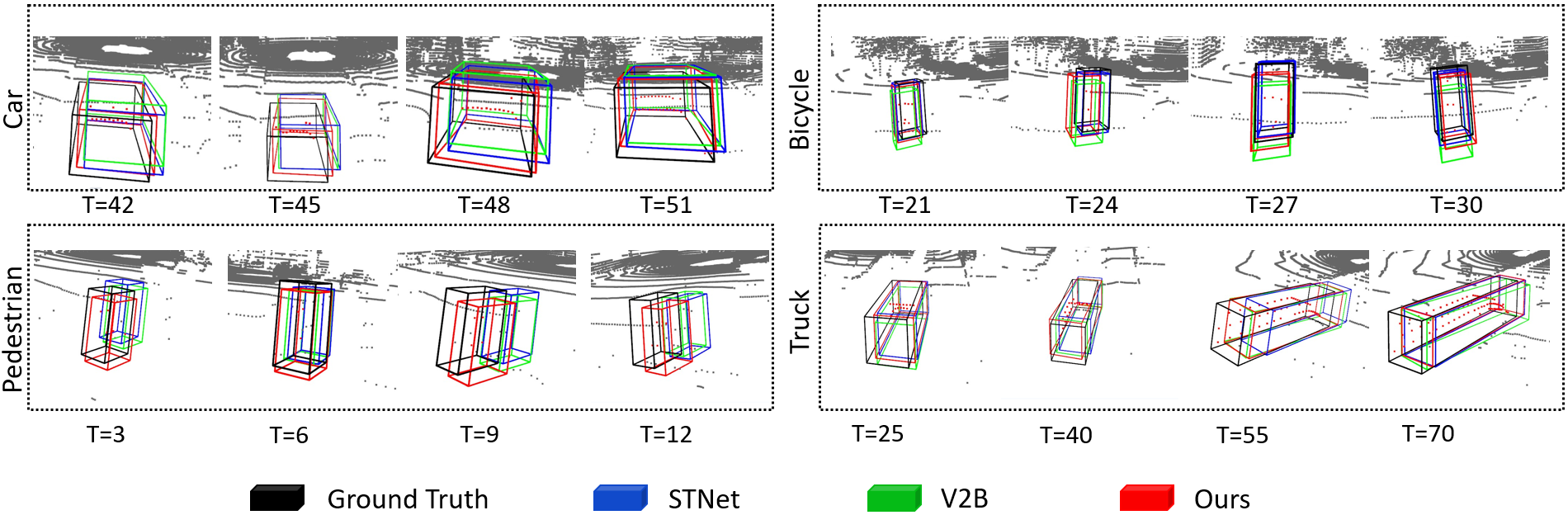}
	\caption{Visualization results of the nuScenes dataset.}
	\label{fig:nuScenes_visual}
\end{figure*}

\noindent\textbf{Results on nuScenes.}  On the nuScenes dataset, we compare our algorithm with SC3D~\cite{SC3D}, P2B~\cite{P2B}, BAT~\cite{BAT}, V2B~\cite{V2B}, STNet~\cite{STNet}, and CXTrack~\cite{CXTrack}. These algorithms are selected because  the same protocol is used to adapt the nuScenes dataset to the 3D SOT task.  The results in Table~\ref{table:result-nuScenes} show that our algorithm achieves the best mean success while CXTrack~\cite{CXTrack} obtains the best mean precision. In general, our tracker is on par with CXTrack on the nuScenes dataset.   Meanwhile, the performance of our tracker in the car category is the best. Since all models evaluated on nuScenes are trained with the KITTI dataset, the results manifest the strong generalization ability of our model. For the pedestrian and truck categories, our tracker ranks second and third, respectively. Some qualitative  results are shown in  Figure~\ref{fig:nuScenes_visual}.

\noindent\textbf{Results on WOD.} We evaluate the performance of our tracker on WOD by comparing with P2B~\cite{P2B}, BAT~\cite{BAT}, V2B~\cite{V2B}, STNet~\cite{STNet}, and CXTrack~\cite{CXTrack}. The selected algorithms use the same approach as ours to suit WOD to the 3D SOT task, and all models evaluated on WOD are also trained with the KITTI dataset.  Results in Table~\ref{table:result-WOD} show that in  the vehicle category, our tracker achieves the best mean performance and the best performances on all three subsets with different difficulties in terms of both success and precision. In the pedestrian category, CXTrack~\cite{CXTrack} performs best while our tracker ranks second.  Compared with nuScenes, the trackers can generalize to WOD better, and the reason should be that both KITTI and WOD use 64-beam LIDARs  while the nuScenes uses 32-beam LIDARs.  


\noindent\textbf{Running speed.}
We  use all test frames of the KITTI car category to report the average inference speed on a single RTX 2080 Ti GPU. Our method runs at 25 FPS while V2B~\cite{V2B} and STNet~\cite{STNet} run at 20 FPS and 19 FPS, respectively.

\subsection{Ablation study}
\label{ablation study}
We comprehensively  validate the effectiveness of each design choice via ablated experiments in this section.   Following~\cite{STNet,BAT,P2B}, all ablation studies are conducted on the car class of the KITTI dataset.  

\begin{table}[htbp]
	\begin{minipage}[t]{0.23\textwidth}
		\centering
		\begin{tabular}{@{\hspace{1mm}}c@{\hspace{1mm}}|@{\hspace{1mm}}c@{\hspace{1mm}}|@{\hspace{1mm}}c@{\hspace{1mm}}}
			\toprule[2pt]
			Stages & Success & Precision  \\
			\midrule
			1 & 72.0 & 83.5  \\
			2 & \textbf{73.6} & \textbf{84.7} \\
			3 & 72.1 & 83.7  \\
			4 & 72.9 & 84.2  \\
			\bottomrule[2pt]
		\end{tabular}
		\caption{Comparison of different numbers of stages.}
		\label{table:stage-choice}
	\end{minipage}
	\hspace{2mm}
	\begin{minipage}[t]{0.23\textwidth}
		\centering
		\begin{tabular}{@{\hspace{1mm}}c@{\hspace{1mm}}|@{\hspace{1mm}}c@{\hspace{1mm}}|@{\hspace{1mm}}c@{\hspace{1mm}}}
			\toprule[2pt]
			Weight ($\alpha$) & Success & Precision  \\
			\midrule
			0 & 72.5 & 83.7  \\ 
			0.1 & \textbf{73.6} & \textbf{84.7}  \\
			0.2 & 71.3 & 83.2 \\
			0.3 & 72.5 & 84.0  \\
			\bottomrule[2pt]
		\end{tabular}
		\caption{Comparison of different weights for deep supervision.}
		\label{table:deep-supervision-weight}
	\end{minipage}
\end{table}

\noindent\textbf{Choice of the number of stages.}  We train our model with different numbers of stages to see how this affects the performance of our tracker.  In Table~\ref{table:stage-choice}, we increase the number of stages from one to four and show the success and precision. It can be observed the two stages setting achieves  the best performance, and we choose two stages as the default setting.

\noindent\textbf{Effectiveness and weight choice of deep supervision.}  We study the effectiveness and find the optimal weight of deep supervision by varying $\alpha$ in Eq.~\ref{eq:finalloss}.  Table~\ref{table:deep-supervision-weight} shows the success and precision for different values of $\alpha$, where $\alpha=0$ corresponds to not adding deep supervision. The results demonstrate that deep supervision can improve the performance and the optimal weight is $\alpha=0.1$. 

\noindent\textbf{Effectiveness of correlations in multiple stages.} To verify the effectiveness of feature correlations in multiple stages, we train models that only carry out feature correlation at the end of the last stage. As template information is not injected into the search region  in the stages before the last one, deep supervision is not suitable for the single correlation approach. For fairness, we remove the deep supervision for the multi-correlation.  The results in Table~\ref{table:single-multi-stage-fusion} show that the multi-correlation strategy can bring about performance gain for all different numbers of stages. 

\begin{table}[htbp]
	\caption{Comparison between multi-stage correlation and single-stage correlation.}
	\label{table:single-multi-stage-fusion}
	\centering
	\begin{tabular}{c|c|c|c|c}
		\toprule[2pt]
		\multirow{2}{*}{Stages}  & \multicolumn{2}{c|}{Single correlation} & \multicolumn{2}{c}{Multi-correlation}  \\
		\cmidrule{2-5}
		& Success & Precision & Success & Precision  \\
		\midrule
		2 & 70.6 & 81.7 & 72.5 & 83.7  \\
		3 & 71.3 & 82.4 & 71.8 & 83.3  \\
		4 & 71.6 & 82.4 & 72.1 & 84.0  \\
		\bottomrule[2pt]
	\end{tabular}
\end{table}

\noindent\textbf{Influence of keeping the template intact.}  We study the influence of keeping the template intact during feature extraction by adding a cross-attention in each stage of the template branch as well, and this cross-attention fuses information from the search region to the template. Results in Table~\ref{table:fusion-direction} show that keeping the template intact (Template $\rightarrow$ Search) is useful for all three numbers of stages.

\begin{table}[htbp]
	\caption{Comparison between one-direction and two-direction fusions. The one-direction fusion keeps the template intact.}
	\label{table:fusion-direction}
	\centering
	\begin{tabular}{c|c|c|c|c}
		\toprule[2pt]
		\multirow{2}{*}{Stages}  & \multicolumn{2}{c|}{Template $\rightarrow$ Search} & \multicolumn{2}{c}{Template $\leftrightarrow$ Search}  \\
		\cmidrule{2-5}
		& Success & Precision & Success & Precision  \\
		\midrule
		2 & 73.6 & 84.7 & 72.4 & 83.8  \\
		3 & 72.1 & 83.7 & 71.9 & 83.1  \\
		4 & 72.9 & 84.2 & 71.5 & 82.6  \\
		\bottomrule[2pt]
	\end{tabular}
\end{table}

\noindent\textbf{Effectiveness of dense connection.}  To verify the effectiveness of dense connection, we first remove the dense connection for both the multi-stage Siamese network and the three convolutions used by the target localization network.  As the results in Table~\ref{table:dense-connection-ablation} show, using dense connection in the multi-stage Siamese network for feature extraction can bring about performance gain, and adding dense connection to the target localization network can improve the performance further. 

\begin{table}[htbp]
	\caption{Results of different dense connection settings.}
	\label{table:dense-connection-ablation}
	\centering
	\begin{tabular}{c|c|c}
		\toprule[2pt]
		Settings & Success & Precision  \\
		\midrule
		No dense & 71.4 & 82.8  \\
		Stages only& 72.5 & 83.4  \\
		Stages+Localization  & \textbf{73.6} & \textbf{84.7}  \\
		\bottomrule[2pt]
	\end{tabular}
\end{table}

\section{Conclusions}
In this paper, we present a Transformer-based multi-correlation Siamese network with multiple stages and dense connection for 3D single object tracking in point clouds.  To enable the feature extraction of the search  region to be aware of the template information while keeping the feature extraction process of the template intact, we propose to inject template information into the search region at each stage of the backbone instead of performing feature correlation at a single point.  In addition, we densely connect all stages to ease the optimization, and dense connection is adopted by the target localization network as well. To further facilitate the optimization process, deep supervision is added to each stage. Experiment results on the popular KITTI, nuScenes, and WOD datasets show that the proposed algorithm achieves promising performance and has strong generalization ability.

\bibliographystyle{IEEEtran}
\bibliography{SOT3D}

\begin{thebibliography}{10}
\providecommand{\url}[1]{#1}
\csname url@rmstyle\endcsname
\providecommand{\newblock}{\relax}
\providecommand{\bibinfo}[2]{#2}
\providecommand\BIBentrySTDinterwordspacing{\spaceskip=0pt\relax}
\providecommand\BIBentryALTinterwordstretchfactor{4}
\providecommand\BIBentryALTinterwordspacing{\spaceskip=\fontdimen2\font plus
\BIBentryALTinterwordstretchfactor\fontdimen3\font minus
  \fontdimen4\font\relax}
\providecommand\BIBforeignlanguage[2]{{%
\expandafter\ifx\csname l@#1\endcsname\relax
\typeout{** WARNING: IEEEtran.bst: No hyphenation pattern has been}%
\typeout{** loaded for the language `#1'. Using the pattern for}%
\typeout{** the default language instead.}%
\else
\language=\csname l@#1\endcsname
\fi
#2}}

\bibitem{luo2018fast}
W.~Luo, B.~Yang, and R.~Urtasun, ``Fast and furious: Real time end-to-end 3d
  detection, tracking and motion forecasting with a single convolutional net,''
  in \emph{IEEE Conference on Computer Vision and Pattern Recognition}, 2018.

\bibitem{cui2022exploiting}
Y.~Cui, J.~Shan, Z.~Gu, Z.~Li, and Z.~Fang, ``Exploiting more information in
  sparse point cloud for 3d single object tracking,'' \emph{IEEE Robotics and
  Automation Letters}, vol.~7, no.~4, pp. 11\,926--11\,933, 2022.

\bibitem{wu2022multi}
Q.~Wu, C.~Sun, and J.~Wang, ``Multi-level structure-enhanced network for 3d
  single object tracking in sparse point clouds,'' \emph{IEEE Robotics and
  Automation Letters}, vol.~8, no.~1, pp. 9--16, 2023.

\bibitem{V2B}
L.~Hui, L.~Wang, M.~Cheng, J.~Xie, and J.~Yang, ``3d siamese voxel-to-bev
  tracker for sparse point clouds,'' in \emph{Advances in Neural Information
  Processing Systems}, 2021.

\bibitem{STNet}
L.~Hui, L.~Wang, L.~Tang, K.~Lan, J.~Xie, and J.~Yang, ``3d siamese transformer
  network for single object tracking on point clouds,'' in \emph{European
  Conference on Computer Vision}, 2022.

\bibitem{BAT}
C.~Zheng, X.~Yan, J.~Gao, W.~Zhao, W.~Zhang, Z.~Li, and S.~Cui, ``Box-aware
  feature enhancement for single object tracking on point clouds,'' in
  \emph{IEEE/CVF International Conference on Computer Vision}, 2021.

\bibitem{mm-track}
C.~Zheng, X.~Yan, H.~Zhang, B.~Wang, S.~Cheng, S.~Cui, and Z.~Li, ``Beyond 3d
  siamese tracking: A motion-centric paradigm for 3d single object tracking in
  point clouds,'' in \emph{IEEE/CVF Conference on Computer Vision and Pattern
  Recognition}, 2022.

\bibitem{bertinetto2016fully}
L.~Bertinetto, J.~Valmadre, J.~F. Henriques, A.~Vedaldi, and P.~H. Torr,
  ``Fully-convolutional siamese networks for object tracking,'' in
  \emph{European Conference on Computer Vision Workshop}, 2016.

\bibitem{li2018high}
B.~Li, J.~Yan, W.~Wu, Z.~Zhu, and X.~Hu, ``High performance visual tracking
  with siamese region proposal network,'' in \emph{IEEE Conference on Computer
  Vision and Pattern Recognition}, 2018.

\bibitem{chen2021transformer}
X.~Chen, B.~Yan, J.~Zhu, D.~Wang, X.~Yang, and H.~Lu, ``Transformer tracking,''
  in \emph{IEEE/CVF Conference on Computer Vision and Pattern Recognition},
  2021.

\bibitem{SC3D}
S.~Giancola, J.~Zarzar, and B.~Ghanem, ``Leveraging shape completion for 3d
  siamese tracking,'' in \emph{IEEE/CVF Conference on Computer Vision and
  Pattern Recognition}, 2019.

\bibitem{P2B}
H.~Qi, C.~Feng, Z.~Cao, F.~Zhao, and Y.~Xiao, ``P2b: Point-to-box network for
  3d object tracking in point clouds,'' in \emph{IEEE/CVF Conference on
  Computer Vision and Pattern Recognition}, 2020.

\bibitem{cui20213dbmvc}
Y.~Cui, Z.~Fang, J.~Shan, Z.~Gu, and S.~Zhou, ``3d object tracking with
  transformer,'' \emph{British Machine Vision Conference}, 2021.

\bibitem{fan2022embracing}
L.~Fan, Z.~Pang, T.~Zhang, Y.-X. Wang, H.~Zhao, F.~Wang, N.~Wang, and Z.~Zhang,
  ``Embracing single stride 3d object detector with sparse transformer,'' in
  \emph{IEEE/CVF Conference on Computer Vision and Pattern Recognition}, 2022.

\bibitem{qi2017pointnet}
C.~R. Qi, H.~Su, K.~Mo, and L.~J. Guibas, ``Pointnet: Deep learning on point
  sets for 3d classification and segmentation,'' in \emph{IEEE Conference on
  Computer Vision and Pattern Recognition}, 2017.

\bibitem{densenet}
G.~Huang, Z.~Liu, L.~Van Der~Maaten, and K.~Q. Weinberger, ``Densely connected
  convolutional networks,'' in \emph{IEEE Conference on Computer Vision and
  Pattern Recognition}, 2017.

\bibitem{kiitidataset}
A.~Geiger, P.~Lenz, and R.~Urtasun, ``Are we ready for autonomous driving? the
  kitti vision benchmark suite,'' in \emph{IEEE Conference on Computer Vision
  and Pattern Recognition}, 2012.

\bibitem{nuscenesdataset}
H.~Caesar, V.~Bankiti, A.~H. Lang, S.~Vora, V.~E. Liong, Q.~Xu, A.~Krishnan,
  Y.~Pan, G.~Baldan, and O.~Beijbom, ``nuscenes: A multimodal dataset for
  autonomous driving,'' in \emph{IEEE/CVF Conference on Computer Vision and
  Pattern Recognition}, 2020.

\bibitem{waymodataset}
P.~Sun, H.~Kretzschmar, X.~Dotiwalla, A.~Chouard, V.~Patnaik, P.~Tsui, J.~Guo,
  Y.~Zhou, Y.~Chai, B.~Caine, \emph{et~al.}, ``Scalability in perception for
  autonomous driving: Waymo open dataset,'' in \emph{IEEE/CVF Conference on
  Computer Vision and Pattern Recognition}, 2020.

\bibitem{zhang2020ocean}
Z.~Zhang, H.~Peng, J.~Fu, B.~Li, and W.~Hu, ``Ocean: Object-aware anchor-free
  tracking,'' in \emph{European Conference on Computer Visio}, 2020.

\bibitem{li2019siamrpn++}
B.~Li, W.~Wu, Q.~Wang, F.~Zhang, J.~Xing, and J.~Yan, ``Siamrpn++: Evolution of
  siamese visual tracking with very deep networks,'' in \emph{IEEE/CVF
  Conference on Computer Vision and Pattern Recognition}, 2019, pp. 4282--4291.

\bibitem{lin2022swintrack}
L.~Lin, H.~Fan, Z.~Zhang, Y.~Xu, and H.~Ling, ``Swintrack: A simple and strong
  baseline for transformer tracking,'' \emph{Advances in Neural Information
  Processing Systems}, 2022.

\bibitem{cui2022mixformer}
Y.~Cui, C.~Jiang, L.~Wang, and G.~Wu, ``Mixformer: End-to-end tracking with
  iterative mixed attention,'' in \emph{IEEE/CVF Conference on Computer Vision
  and Pattern Recognition}, 2022.

\bibitem{vaswani2017attention}
A.~Vaswani, N.~Shazeer, N.~Parmar, J.~Uszkoreit, L.~Jones, A.~N. Gomez,
  {\L}.~Kaiser, and I.~Polosukhin, ``Attention is all you need,'' 2017.

\bibitem{liu2021swin}
Z.~Liu, Y.~Lin, Y.~Cao, H.~Hu, Y.~Wei, Z.~Zhang, S.~Lin, and B.~Guo, ``Swin
  transformer: Hierarchical vision transformer using shifted windows,'' in
  \emph{IEEE/CVF International Conference on Computer Vision}, 2021.

\bibitem{pieropan2015robust}
A.~Pieropan, N.~Bergstr{\"o}m, M.~Ishikawa, and H.~Kjellstr{\"o}m, ``Robust 3d
  tracking of unknown objects,'' in \emph{IEEE International Conference on
  Robotics and Automation}, 2015.

\bibitem{kart2019object}
U.~Kart, A.~Lukezic, M.~Kristan, J.-K. Kamarainen, and J.~Matas, ``Object
  tracking by reconstruction with view-specific discriminative correlation
  filters,'' in \emph{IEEE/CVF Conference on Computer Vision and Pattern
  Recognition}, 2019.

\bibitem{song2013tracking}
S.~Song and J.~Xiao, ``Tracking revisited using rgbd camera: Unified benchmark
  and baselines,'' in \emph{IEEE International Conference on Computer Vision},
  2013.

\bibitem{shan2021ptt}
J.~Shan, S.~Zhou, Z.~Fang, and Y.~Cui, ``Ptt: Point-track-transformer module
  for 3d single object tracking in point clouds,'' in \emph{IEEE/RSJ
  International Conference on Intelligent Robots and Systems}, 2021.

\bibitem{wang2021mlvsnet}
Z.~Wang, Q.~Xie, Y.-K. Lai, J.~Wu, K.~Long, and J.~Wang, ``Mlvsnet: Multi-level
  voting siamese network for 3d visual tracking,'' in \emph{IEEE/CVF
  International Conference on Computer Vision}.

\bibitem{zhou2022pttr}
C.~Zhou, Z.~Luo, Y.~Luo, T.~Liu, L.~Pan, Z.~Cai, H.~Zhao, and S.~Lu, ``Pttr:
  Relational 3d point cloud object tracking with transformer,'' in
  \emph{IEEE/CVF Conference on Computer Vision and Pattern Recognition}, 2022.

\bibitem{li2022deepfusion}
Y.~Li, A.~W. Yu, T.~Meng, B.~Caine, J.~Ngiam, D.~Peng, J.~Shen, Y.~Lu, D.~Zhou,
  Q.~V. Le, \emph{et~al.}, ``Deepfusion: Lidar-camera deep fusion for
  multi-modal 3d object detection,'' in \emph{IEEE/CVF Conference on Computer
  Vision and Pattern Recognition}, 2022.

\bibitem{bai2022transfusion}
X.~Bai, Z.~Hu, X.~Zhu, Q.~Huang, Y.~Chen, H.~Fu, and C.-L. Tai, ``Transfusion:
  Robust lidar-camera fusion for 3d object detection with transformers,'' in
  \emph{IEEE/CVF Conference on Computer Vision and Pattern Recognition}, 2022.

\bibitem{yang2022deepinteraction}
Z.~Yang, J.~Chen, Z.~Miao, W.~Li, X.~Zhu, and L.~Zhang, ``Deepinteraction: 3d
  object detection via modality interaction,'' in \emph{Advances in Neural
  Information Processing Systems}, 2022.

\bibitem{autoalignchen2022}
Z.~Chen, Z.~Li, S.~Zhang, L.~Fang, Q.~Jiang, F.~Zhao, B.~Zhou, and H.~Zhao,
  ``Autoalign: Pixel-instance feature aggregation for multi-modal 3d object
  detection,'' in \emph{International Joint Conference on Artificial
  Intelligence}, 2022.

\bibitem{chen2022autoalignv2}
Z.~Chen, Z.~Li, S.~Zhang, L.~Fang, Q.~Jiang, and F.~Zhao, ``Autoalignv2:
  Deformable feature aggregation for dynamic multi-modal 3d object detection,''
  in \emph{European Conference on Computer Vision}, 2022.

\bibitem{tong2023multi}
G.~Tong, Z.~Li, H.~Peng, and Y.~Wang, ``Multi-source features fusion single
  stage 3d object detection with transformer,'' \emph{IEEE Robotics and
  Automation Letters}, vol.~8, no.~4, pp. 2062--2069, 2023.

\bibitem{pointpillars}
A.~H. Lang, S.~Vora, H.~Caesar, L.~Zhou, J.~Yang, and O.~Beijbom,
  ``Pointpillars: Fast encoders for object detection from point clouds,'' in
  \emph{IEEE/CVF Conference on Computer Vision and Pattern Recognition}, 2019.

\bibitem{zhou2020end}
Y.~Zhou, P.~Sun, Y.~Zhang, D.~Anguelov, J.~Gao, T.~Ouyang, J.~Guo, J.~Ngiam,
  and V.~Vasudevan, ``End-to-end multi-view fusion for 3d object detection in
  lidar point clouds,'' in \emph{Conference on Robot Learning}, 2020.

\bibitem{katharopoulos2020transformers}
A.~Katharopoulos, A.~Vyas, N.~Pappas, and F.~Fleuret, ``Transformers are rnns:
  Fast autoregressive transformers with linear attention,'' in
  \emph{International Conference on Machine Learning}, 2020.

\bibitem{lin2017focal}
T.-Y. Lin, P.~Goyal, R.~Girshick, K.~He, and P.~Doll{\'a}r, ``Focal loss for
  dense object detection,'' in \emph{IEEE International Conference on Computer
  Vision}, 2017.

\bibitem{xie2015holistically}
S.~Xie and Z.~Tu, ``Holistically-nested edge detection,'' in \emph{IEEE
  International Conference on Computer Vision}, 2015.

\bibitem{CXTrack}
T.-X. Xu, Y.-C. Guo, Y.-K. Lai, and S.-H. Zhang, ``Cxtrack: Improving 3d point
  cloud tracking with contextual information,'' in \emph{IEEE Conference on
  Computer Vision and Pattern Recognition}, 2023.

\bibitem{nie2023osp2b}
J.~Nie, Z.~He, Y.~Yang, Z.~Bao, M.~Gao, and J.~Zhang, ``Osp2b: One-stage
  point-to-box network for 3d siamese tracking,'' in \emph{International Joint
  Conferences on Artificial Intelligence}, 2023.

\bibitem{LidarSOT}
Z.~Pang, Z.~Li, and N.~Wang, ``Model-free vehicle tracking and state estimation
  in point cloud sequences,'' in \emph{IEEE/RSJ International Conference on
  Intelligent Robots and Systems}.\hskip 1em plus 0.5em minus 0.4em\relax IEEE,
  2021.

\bibitem{wu2013online}
Y.~Wu, J.~Lim, and M.-H. Yang, ``Online object tracking: A benchmark,'' in
  \emph{IEEE conference on Computer Vision and Pattern Recognition}, 2013.

\bibitem{mmdet3d2020}
M.~Contributors, ``{MMDetection3D: OpenMMLab} next-generation platform for
  general {3D} object detection,''
  \url{https://github.com/open-mmlab/mmdetection3d}, 2020.

\bibitem{adam}
D.~P. Kingma and J.~Ba, ``Adam: A method for stochastic optimization,''
  \emph{arXiv preprint arXiv:1412.6980}, 2014.

\end{thebibliography}

\end{document}